%% file: main.tex
\documentclass[conference]{IEEEtran}
\IEEEoverridecommandlockouts

\usepackage{cite}
\usepackage{amsmath,amssymb,amsfonts}
\usepackage{algorithmic}
\usepackage{algorithm,algorithmic}
\usepackage{graphicx}
\usepackage{textcomp}
\usepackage{xcolor}
\usepackage{multirow}
\usepackage[export]{adjustbox}
\def\BibTeX{{\rm B\kern-.05em{\sc i\kern-.025em b}\kern-.08em
    T\kern-.1667em\lower.7ex\hbox{E}\kern-.125emX}}
\DeclareMathOperator{\argmax}{argmax}
\usepackage{pdfpages}
\usepackage{balance}

\begin{document}

\title{Label Correction for Road Segmentation Using Road-side Cameras\\
}

\author{\IEEEauthorblockN{1\textsuperscript{st} Henrik Toikka}
\IEEEauthorblockA{
\textit{Energy and Mechanical Engineering} \\
\textit{Aalto University}\\
Espoo, Finland \\
henrik.toikka@aalto.fi}
\and
\IEEEauthorblockN{2\textsuperscript{nd} Eerik Alamikkotervo}
\IEEEauthorblockA{
\textit{Energy and Mechanical Engineering} \\
\textit{Aalto University}\\
Espoo, Finland \\
eerik.alamikkotervo@aalto.fi}
\and
\IEEEauthorblockN{3\textsuperscript{rd} Risto Ojala}
\IEEEauthorblockA{
\textit{\shortstack[l]{Energy and Mechanical Engineering}} \\
\textit{Aalto University}\\
Espoo, Finland \\
risto.j.ojala@aalto.fi}
}

\maketitle

\begin{abstract}

Reliable road segmentation in all weather conditions is critical for intelligent transportation applications, autonomous vehicles and advanced driver's assistance systems. For robust performance, all weather conditions should be included in the training data of deep learning-based perception models. However, collecting and annotating such a dataset requires extensive resources. In this paper, existing roadside camera infrastructure is utilized for collecting road data in varying weather conditions automatically. Additionally, a novel semi-automatic annotation method for roadside cameras is proposed. For each camera, only one frame is labeled manually and then the label is transferred to other frames of that camera feed. The small camera movements between frames are compensated using frequency domain image registration. The proposed method is validated with roadside camera data collected from 927 cameras across Finland over 4 month time period during winter. Training on the semi-automatically labeled data boosted the segmentation performance of several deep learning segmentation models. Testing was carried out on two different datasets to evaluate the robustness of the resulting models. These datasets were an in-domain roadside camera dataset and out-of-domain dataset captured with a vehicle on-board camera.  

\end{abstract}


\section{Introduction}

\IEEEPARstart{R}{oad} segmentation from a camera view is a core task in many intelligent transportation applications.
Self-driving cars and advanced driver assistance systems (ADAS) rely on precise road segmentation to identify the safely drivable area. Different traffic monitoring systems based on roadside cameras may also incorporate road segmentation for scene understanding of the traffic environment. Currently, deep learning models provide state-of-the-art performance in perception tasks, including road segmentation. However, these models are only reliable in the domain of their training data, meaning all possible road conditions should be included in the training data for reliable performance. The weather is one of the most important factors affecting the visual condition of the road. Varying amounts of water, ice, snow, fog, and sunlight can change the road's appearance drastically~\cite{shaik2024idd,sakaridis2021acdc}. It is challenging to capture a road dataset including all possible weather conditions, and especially to annotate it. 

In most countries, there is an existing infrastructure of roadside cameras for monitoring the traffic, weather, and road conditions. The roadside cameras are usually stationary and publish a continuous feed of images year-round, capturing a variety of weather and lighting conditions. If the road is manually annotated to one roadside camera frame the same label can be transferred to other frames from that camera with no extra effort assuming the perspective does not change. However, wind, temperature changes, and vibrations can change the orientation of the camera over long time periods, compromising the accuracy of the transferred labels. The small perspective changes could be potentially approximated using image registration techniques. Image registration techniques have been proven effective in multiple domains~\cite{marcenaro2001image,markel1972sift,tong2019image}, but their suitability for roadside camera feeds has not been explored. 

In this paper, we propose a semi-automatic road segmentation dataset generation method leveraging existing roadside camera infrastructure and accurate label transfer with image registration. We validate the usability of our approach by training road segmentation models on the resulting data. Models are tested on both roadside and on-board camera data to demonstrate the potential of acquiring rich and diverse road appearance data from roadside cameras.  We are the first to propose and demonstrate methods for transferring a single manual road annotation to all other frames of a roadside camera. We are also the first to propose using road side cameras to train road segmentation models for on-board vehicle use. 
The models trained with our proposed label transfer strategy outperformed the baseline models tested for comparison.

\section{Related Work}

\subsection{Roadside Cameras in Intelligent Transport Systems}

Roadside cameras are common equipment in road infrastructure, providing an image feed of the road from a fixed position. They are widely available, and offer a lucrative source of diverse and rich data source for different perception tasks~\cite{ye2022rope3d, cress2024tumtraf}. Traffic monitoring and surveillance have traditionally been the most common applications~\cite{wu2006traffic, douxchamps2006high, dubska2014fully, schoepflin2003dynamic}, with the perception task typically featuring detection and localization of road users~\cite{zou2022real, zhang2024evaluating}. Roadside cameras have also been utilized for monitoring of the road condition in varying weather~\cite{ojala2024road, carrillo2020integration}. However, research on using roadside cameras for road segmentation is limited. Classical computer vision techniques such as background segmentation have been used to segment road areas from roadside camera views~\cite{fang2015automatic}, yet methods based on recent deep learning approaches have not been investigated.

\subsection{Image registration}

Image registration is a potential tool for determining and correcting changes in a roadside camera view. Image registration is the process of aligning images taken from different viewpoints of the same scene. Image registration techniques can be divided into two main categories: feature and dense-matching-based registration~\cite{marcenaro2001image}. Feature-based image registration aims to find the transform by matching image features between frames~\cite{markel1972sift}. However, feature-based techniques are not suited for roadside camera use case as the selected features can be easily corrupted by weather and illumination changes. On the other hand, dense-matching algorithms consider all pixels of the images to find optimal alignment, making them more robust to noise.  Dense-matching methods typically find the optimal transform by computing the phase correlation of all pixels in the frequency domain~\cite{tong2019image}. The frequency domain methods only support translation, scaling and rotation transforms, but these are sufficient for a roadside camera usecase with limited changes in the view. 



\subsection{Road segmentation in challenging conditions}

There have been efforts to create annotated road segmentation datasets from adverse weather~\cite{sakaridis2021acdc, kurup2023winter,diaz2022ithaca365,shaik2024idd,vachmanus2020semantic}, which allows development of supervised road segmentation models. However, the number of annotated images included is limited as manual annotation work is extremely time-consuming. Thus, all weather conditions can't be included in the data leading to unreliable predictions in out-of-domain scenarios. 

Trajectory-based road segmentation methods can learn to segment the road without manual annotations using the traversed route as the only supervision~\cite{alamikkotervo2024tadap,alamikkotervo2024trajectory,seo2023learning,schmid2022self,jung2024v}, allowing easy adaptation to varying driving conditions. However, it is challenging to learn to detect areas not included in the traversed route without explicit supervision. 

Typical deep learning models for the segmentation task include convolutional neural network models such as DeepLabv3~\cite{chen2017rethinking}. Recently, transformer models have gained popularity in segmentation tasks. This has been expedited with the rise of foundation models, such as DINOv2~\cite{oquab2023dinov2}, which have been extensively pretrained on vast datasets. Foundation models can provide improved performance in out-of-distribution scenarios~\cite{wei2024stronger}.


\section{Methods}

The proposed label transfer method includes two steps. First the image registration transforms between the frames are computed using Fourier-Mellin. The image registration process is presented in section~\ref{sec:image_registration}. If the changes between frames are large, the registration process fails, meaning that it is not feasible to directly compute transform from the reference frame to each of the target frames. Instead, we chain multiple transforms to find an optimal path from the reference frame to each of the target frames. The path finding method is presented in Section \ref{sec:path_finding}. 

\subsection{Image registration}
\label{sec:image_registration}

The goal is to move the manual annotation from the reference frame to other frames of feed. 
However, the roadside camera can have small movements between frames, which causes error if not compensated. Image registration can align two images taken from different viewpoints (Fig.~\ref{fig:imgreg}), allowing us to correct the labels position. 

In this paper, we use dense frequency based Fourier-Mellin transform~\cite{chen1994symmetric} as it is robust to noise caused by weather effects and lighting changes. It is commonly used for registering satellite images~\cite{tong2019image}. Fourier-Mellin only supports Euclidian transforms: translation, scaling and rotation. Shear effects, lens distortions, and other non-Euclidean disturbances are assumed to be negligible. As the camera location is quite far from the road and the movements are small this assumption can be made for road side cameras.

In the Fourier-Mellin image registration process, the source and target images are first preprocessed by converting them to grayscale, applying a Hanning window to reduce edge effects, and converted to frequency domain with Fast Fourier Transform (FFT).
A high-pass filter is applied to reduce high frequency noise and the images are then converted to log-polar coordinates.
After preprocessing, the scale and rotation between two images can be calculated with phase correlation by finding the coordinates of the peak response.
After correcting the reference image for rotation and scale, another phase correlation is used in the pixel domain images to compute the pixel translation between the images.
In this paper, we use the implementation of phase correlation and discrete fourier transforms provided by OpenCV.
Additionally, OpenCV's implementation of phase correlation provides the normalized signal power from a 5x5 centroid around the peak between 0 and 1, which can used as an analogue for the registration quality.
This measure is later used for finding optimal transformation chains when registering image feeds.

\subsection{Finding optimal path of transformations}
\label{sec:path_finding}

Drastic weather and lightning condition changes in road side camera data can cause the Fourier-Mellin registration to fail. The number of failures can be significantly lowered by finding a chain of transforms instead trying to directly go from the reference frame to the target frame. Assuming each transformation has sub-pixel accuracy, the maximum error of the combined transform will be at maximum the transformation chain length in pixels. 

The Fourier-Mellin registration returns a response value along with the transformation. We define the optimal chain of transformations as the chain with the highest product of transformation responses. 
Let $P(k,i_\text{ref})$ be the set of all possible paths from target image $k$ to reference image $i_\text{ref}$. For a path $p\in P(k,i_\text{ref})$, let $p=(k,i_1,i_2,\dots,i_\text{ref})$, where subsequent indices $p_m$ and $p_{m+1}$ represent a transition from image $m$ to $m+1$. Let the response of the Fourier-Mellin transform be $r(i,j)$, where $i$ and $j$ are image indices. The score of the path is calculated by multiplying the response of each transition. The optimal path $p^*(k,i_\text{ref})$ is then defined as:

\begin{align}
    p^*(k,i_\text{ref})&=\underset{p\in P(k,i_\text{ref})}{\argmax}\prod_{m=1}^{|p|-1}r\left(p_m,p_{m+1}\right)\label{eq_optimal_path}
\end{align}


{\setlength\tabcolsep{2 pt}
\begin{figure}
\centering
\begin{tabular}{c|c|c}
\shortstack[l]{Frame with\\manual label} & \shortstack[l]{Another frame\\w/ direct reuse} & \shortstack[l]{Corrected label w/\\image registration} \\
    \includegraphics[width=0.15\textwidth]{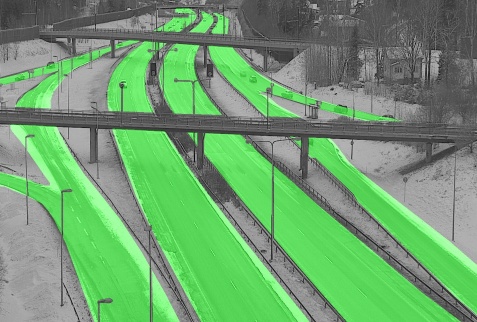} & 
    \includegraphics[width=0.15\textwidth]{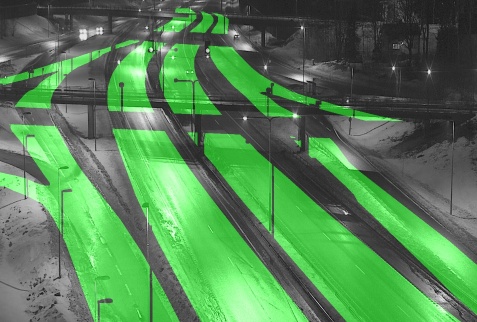} & 
    \includegraphics[width=0.15\textwidth]{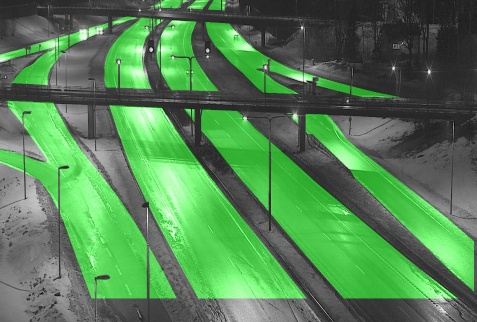} \\
    \includegraphics[width=0.15\textwidth]{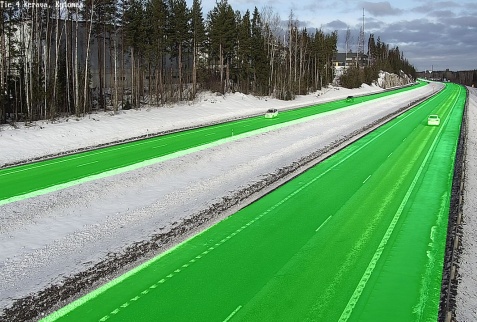} & 
    \includegraphics[width=0.15\textwidth]{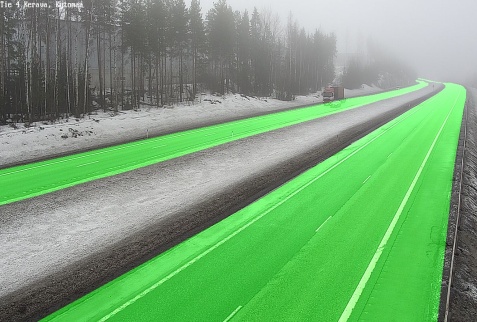} & 
    \includegraphics[width=0.15\textwidth]{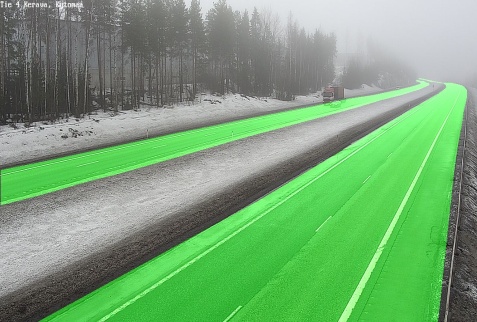} \\
    \includegraphics[width=0.15\textwidth]{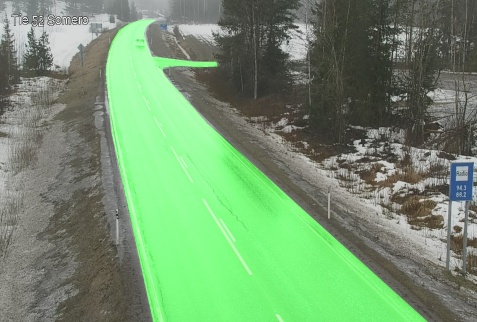} & 
    \includegraphics[width=0.15\textwidth]{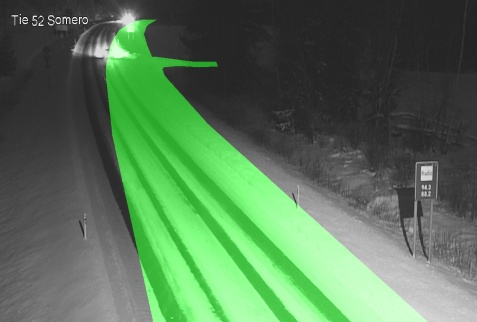} & 
    \includegraphics[width=0.15\textwidth]{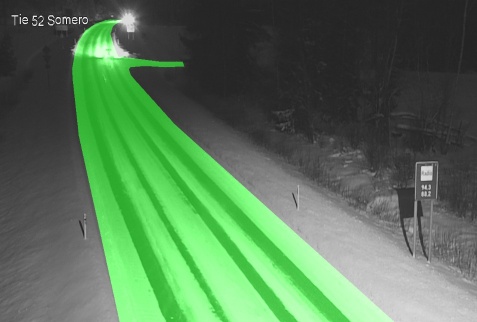} \\
\end{tabular}
\caption{Original manual labels are shown on the left column. Center and right columns display another frame from the same camera with a directly reused label and a corrected label, respectively.}
\label{fig:imgreg}
\end{figure}
}


Computation of all transforms between $N$ images would require $N^2$ image registration operations. 
To decrease computational load, all transformations between frames were not computed when searching for the optimal transform chain. 

The frames of each feed where divided in to batches of 24 images based on the timestamp. All transforms between these images were computed. However, only some of the transforms were computed between images from different batches. The proportion of computed transforms was defined by $\gamma^d$ where $d$ is the number of batches between the images and $\gamma=1/1.35$ is a decay factor. The further away the batches were from each other, the smaller proportion of transforms was computed. If the number of batches $d$ between images was more than 8, no transforms were computed. 

Regardless of the lower number of computed transforms, there are still a high number of possible paths between any given pair of images, while the computational load is significantly decreased. Even when using chaining, the registration process may still fail due to extreme weather conditions, low lighting or corrupted camera frames. If the product of responses for the optimal transform chain \eqref{eq_optimal_path} is below a threshold of 0.45, the registration is considered to be failed and the frame is filtered out from the dataset.


\subsection{Datasets}

\subsubsection{Roadside camera dataset}

The proposed method was validated with a dataset collected from 1025 roadside cameras across Finland from December 2023 to March 2024. Camera feeds with artificial privacy masks were excluded from the analysis, resulting in a total of 927 camera feeds. The cameras were geographically located across the whole country for a maximum variation of locations and weather conditions. During the collection period, a new image was requested from each camera every 20 minutes through the Digitraffic API, resulting in roughly 7000 frames per camera. Out of this data, approximately 10\% was randomly sampled to form the final dataset. The dataset contains various weather conditions and perspectives. 


Out of the total 927 camera feeds, 661 are assigned to training split, 102 to validation split and 206 to test split. 
A randomly chosen frame from each feed is manually labeled.
The test and validation sets remain identical across experiments, but three different training sets are created. 

\begin{itemize}
    \item \textbf{Baseline}. Includes only the manually labeled frame from each feed (661 images). 
    \item \textbf{Reuse}. The manually labeled frame is transferred to all other frames in that feed without correction. Camera movement between frames can cause error to the transferred labels. (493 411 images) 
    \item \textbf{Corrected Reuse}. The manually labeled frame is transferred to all other frames in that feed using image registration to correct camera movements between frames. If registration fails the frame is filtered out. (164 128 images)
\end{itemize}

\subsubsection{Dashcam dataset}

We evaluate the model's capability to generalize from roadside camera training to dashcam perspective using 400 test images from suburban driving in winter and 400 test images from countryside driving in winter. The dataset has been previously used in \cite{alamikkotervo2024trajectory}. In suburban driving the road is partially covered in snow and in countryside driving fully covered in snow. Dashcam data is not used for training. Testing on the dashcam demonstrates the generalizability of models trained on the roadside camera data.

\subsection{Validation}

\begin{figure}
    \centering
    \includegraphics[width=\linewidth]{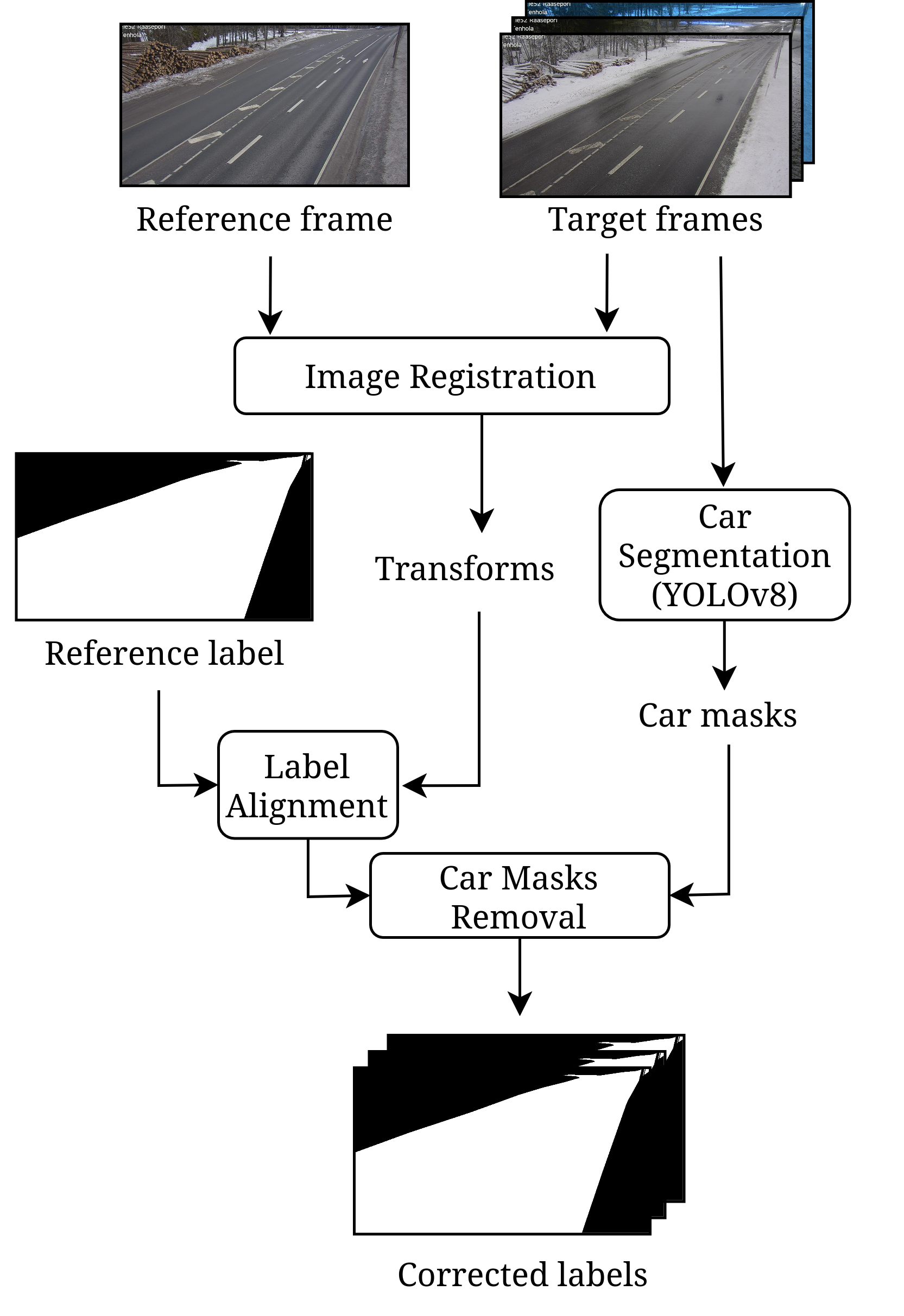}
    \caption{Label correction architecture. Labels are aligned based on image registration, and cars possibly present on the road are segmented out of the label.}
    \label{fig:architecture}
\end{figure}

We trained three different models with the baseline dataset, the reuse dataset and the corrected reuse dataset. 

\begin{enumerate}
    \item \textbf{Deeplabv3} with Resnet50 backbone
    \item \textbf{Dinov2 + Seghead}. Frozen pre-trained Dinov2 used as backbone. Segmentation head with two deconvolution layers trained on top of the frozen features. 
    \item \textbf{Dinov2 + Linear probe}. Frozen pre-trained Dinov2 used as backbone. Linear probe trained on top of the frozen features. 
\end{enumerate}

Each model was trained using a learning rate of 1e-4 and a batch size of 32 until the validation score converges. 
During training images were resized to 644x644 with random crop and during testing images were resized to 644x644 with center crop. 
Each model was evaluated on the roadside and dashcam test set, both unseen during training. 

\section{Results}

Intersection over Union (IoU), Precision (PRE), Recall (REC) and F1-score (F1) are reported for each model. Results for the roadside camera and dashcam datasets are presented in Table \ref{tab:weathercam_table} and Table \ref{tab:dashcam_table}, respectively. On both test sets, highest IoU values are reached by models trained with our proposed Corrected Reuse training set. The highest IoU value on the roadside camera test set was 93.50 reached by DeepLabv3. On the dashcam test set, the highest IoU was 95.35 reached by Dinov2 + Seghead. Examples of predictions by these models on the dashcam dataset are presented in Fig.~\ref{fig:dashcam_visualisation} and on the roadside camera dataset in Fig.~\ref{fig:weathercam_visualisation}. 

{\setlength\tabcolsep{2 pt}
\begin{figure*}
\centering
\begin{tabular}{c|cc|cc|}
& \multicolumn{2}{|c|}{Baseline} & \multicolumn{2}{|c|}{Corrected Reuse} \\
Original image & DeepLabv3 & DinoV2 + Seghead & DeepLabv3 & DinoV2 + Seghead \\
 \includegraphics[width=0.18\textwidth]{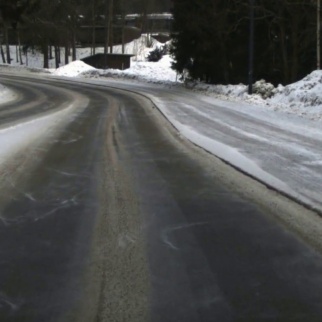} & 
  \includegraphics[width=0.18\textwidth]{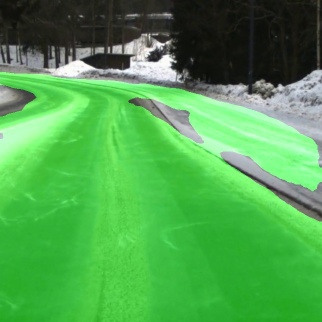} &
   \includegraphics[width=0.18\textwidth]{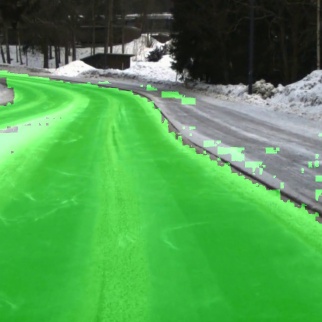} &
    \includegraphics[width=0.18\textwidth]{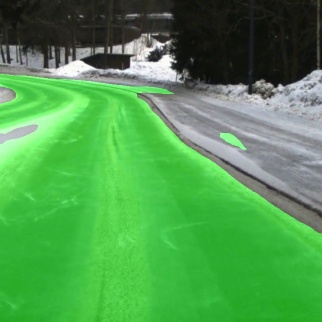} &
     \includegraphics[width=0.18\textwidth]{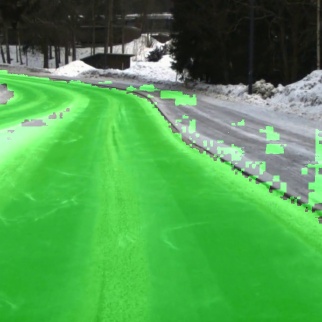} \\

 \includegraphics[width=0.18\textwidth]{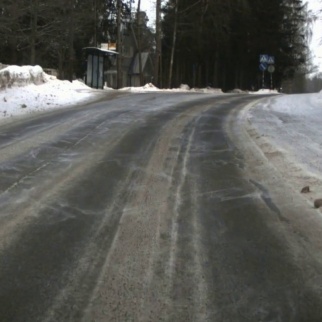} & 
  \includegraphics[width=0.18\textwidth]{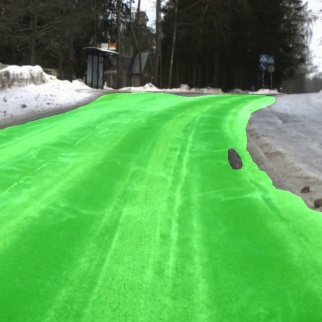} &
   \includegraphics[width=0.18\textwidth]{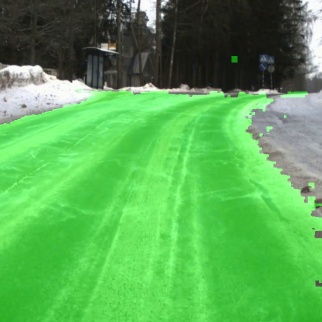} &
    \includegraphics[width=0.18\textwidth]{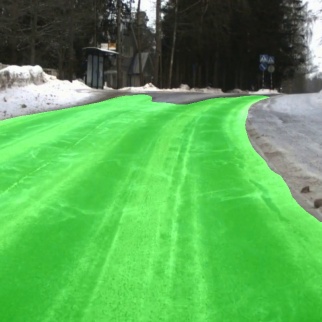} &
     \includegraphics[width=0.18\textwidth]{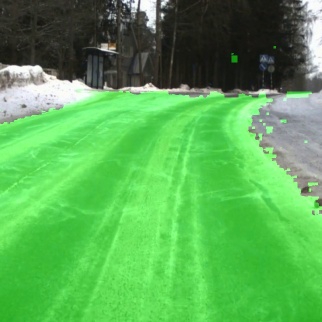} \\

 \includegraphics[width=0.18\textwidth]{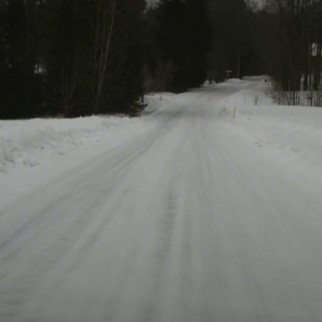} & 
  \includegraphics[width=0.18\textwidth]{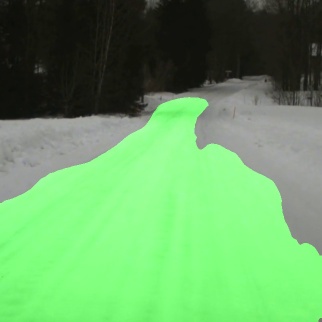} &
   \includegraphics[width=0.18\textwidth]{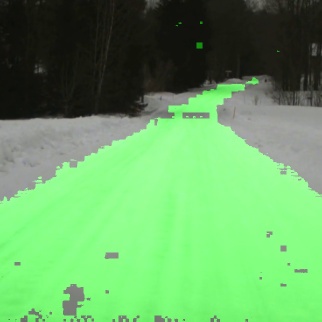} &
    \includegraphics[width=0.18\textwidth]{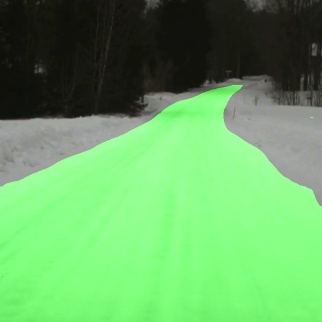} &
     \includegraphics[width=0.18\textwidth]{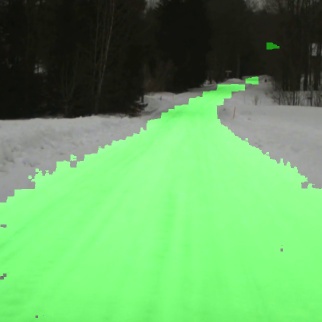} \\

\end{tabular}
\caption{Visualisation of results on the Dashcam dataset.}
\label{fig:dashcam_visualisation}
\end{figure*}
}

{\setlength\tabcolsep{2 pt}
\begin{figure*}
\centering
\begin{tabular}{c|cc|cc|}
& \multicolumn{2}{|c|}{Baseline} & \multicolumn{2}{|c|}{Corrected Reuse} \\
Original image & DeepLabv3 & DinoV2 + Seghead & DeepLabv3 & DinoV2 + Seghead \\
 \includegraphics[width=0.18\textwidth]{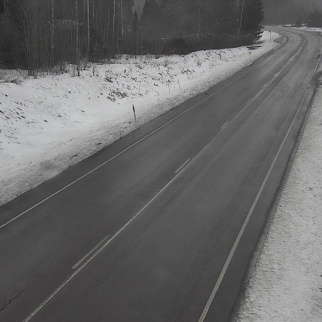} & 
    \includegraphics[width=0.18\textwidth]{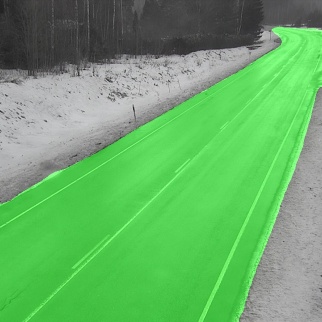} & 
    \includegraphics[width=0.18\textwidth]{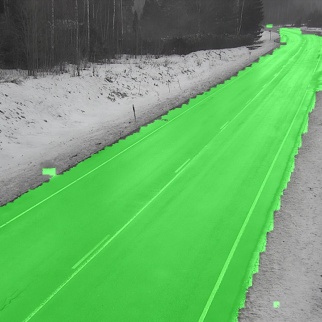} & 
    \includegraphics[width=0.18\textwidth]{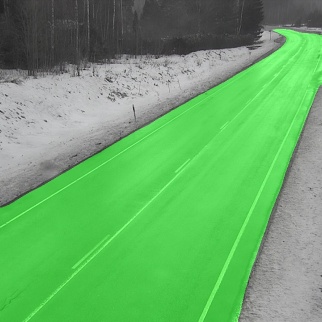} & 
    \includegraphics[width=0.18\textwidth]{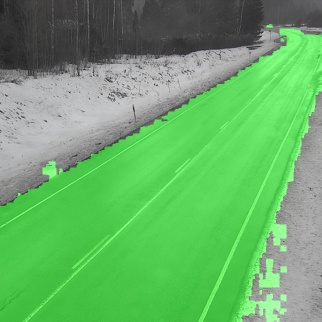} \\
    
 \includegraphics[width=0.18\textwidth]{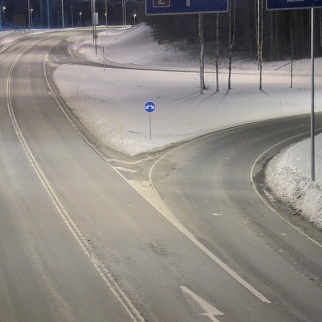} & 
    \includegraphics[width=0.18\textwidth]{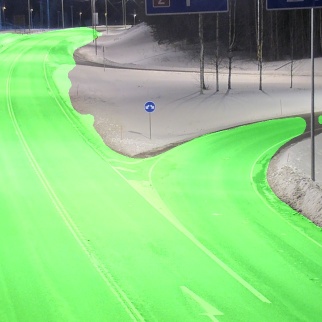} & 
    \includegraphics[width=0.18\textwidth]{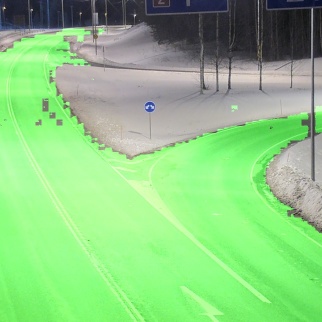} & 
    \includegraphics[width=0.18\textwidth]{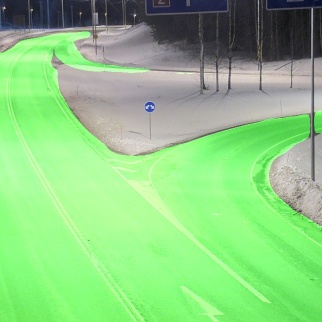} & 
    \includegraphics[width=0.18\textwidth]{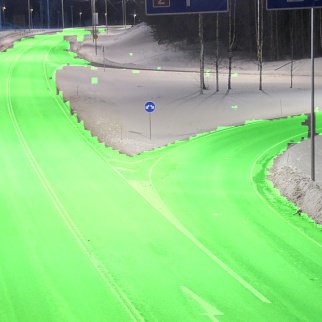} \\
    
 \includegraphics[width=0.18\textwidth]{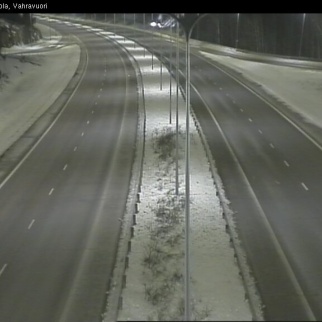} & 
    \includegraphics[width=0.18\textwidth]{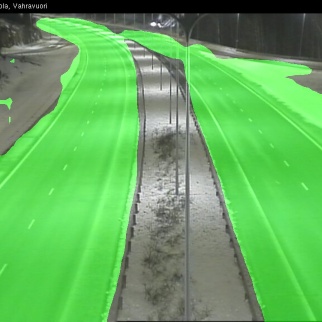} & 
    \includegraphics[width=0.18\textwidth]{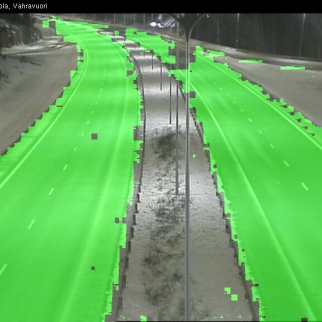} & 
    \includegraphics[width=0.18\textwidth]{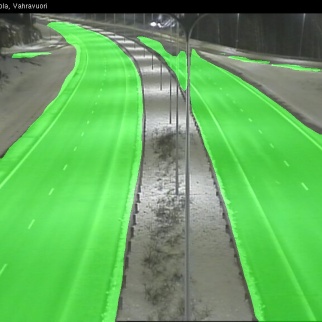} & 
    \includegraphics[width=0.18\textwidth]{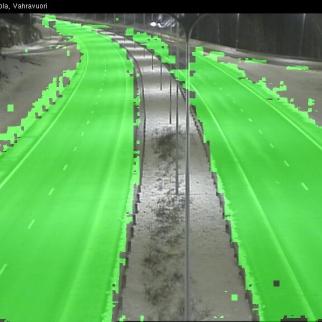} \\

\end{tabular}
\caption{Visualization of results on the roadside camera dataset.}
\label{fig:weathercam_visualisation}
\end{figure*}
}

\begin{table}[ht]
    \centering
    \caption{Test results on the roadside camera dataset. Best IoU-value of each model is bolded.}
    \begin{tabular}{l l c c c c}
        Train data & Model & IoU & PRE & REC & F1\\
        \hline
        \multirow{3}{*}{\shortstack{Baseline}}  & DeepLabv3 & 90.97 & 90.58 & 85.29 & 87.52  \\
        
        & Dinov2 + Seghead & 90.25 & 89.28 & 90.57 & 89.81 \\

        & Dinov2 + Linear probe & 87.61 & 87.88 & 89.13 & 88.29\\

        \hline
         
        \multirow{3}{*}{\shortstack{Reuse}}  & DeepLabv3 & 92.69 & 90.48 & 87.10 & 88.37 \\ 

        & Dinov2 + Seghead & \bf{91.62} & 90.65 & 88.47 & 89.43 \\

        & Dinov2 + Linear probe & 88.90 & 89.55 & 87.95 & 88.60  \\

        \hline

        \multirow{3}{*}{\shortstack[l]{Corrected \\Reuse}}
        & DeepLabv3 & \bf{93.50} & 89.58 & 89.42 & 89.03\\
        
        & Dinov2 + Seghead & 91.61 & 90.61 & 88.38 & 89.32\\
        
        & Dinov2 + Linear probe & \bf{89.91} & 88.89 & 88.99 & 88.80 \\

        \hline
    \end{tabular}
    
    \label{tab:weathercam_table}
\end{table}

\begin{table}[ht]
    \centering
    \caption{Test results on the DashCam dataset. Best IoU-value of each model is bolded.}
    \begin{tabular}{l l c c c c}
        Train data & Model & IoU & PRE & REC & F1\\
        \hline
         \multirow{3}{*}{\shortstack{Baseline}} & DeepLabv3 & 87.83 & 89.18 & 97.76 & 92.58 \\

        & Dinov2 + Seghead & 94.73 & 96.52 & 98.72 & 97.58 \\

        & Dinov2 + Linear probe & \bf{93.62} & 96.15 & 97.88 & 96.96 \\

        \hline
         
         \multirow{3}{*}{\shortstack{Reuse}} & DeepLabv3 & 90.56 & 91.99 & 97.88 & 94.01\\

        & Dinov2 + Seghead & 95.08 & 98.05 & 97.59 & 97.78\\

        & Dinov2 + Linear probe & 92.96 & 97.68 & 96.82 & 97.19\\

        \hline

        \multirow{3}{*}{\shortstack[l]{Corrected \\ Reuse}} & DeepLabv3 & \textbf{94.72} & 96.45 & 98.05 & 97.12 \\

         & Dinov2 + Seghead & \textbf{95.35} & 97.85 & 97.89 & 97.82 \\

         & Dinov2 + Linear probe & 93.24 & 96.54 & 97.68 & 97.04\\

        \hline
        
    \end{tabular}
    
    \label{tab:dashcam_table}
\end{table}

\section{Discussion}

In this paper, we proposed a novel semiautomatic road annotation method for road side cameras, where each manual annotation is transferred to all other frames in the camera feed. Small camera movements between frames are estimated with image registration. The proposed method allows creation of large road segmentation dataset that include varying weather conditions with very little manual work. 

The Deeplabv3 model achieves a clear performance increase with the proposed label reuse, while the Dinov2 based models only see small performance gains. As the Dinov2 backbone is frozen during training, there are few trainable parameters and the models can't properly utilize the additional data. On the other hand, the Dinov2 backbone has already been pre-trained with large amounts of data, lowering the added value of extra training. Label reuse seems to be most valuable when training larger models from scratch. 

In favorable conditions image registration finds accurate alignment between frames, but drastic changes in weather and lighting conditions often cause image registration to fail. To avoid large changes between frames, we searched a path through multiple frames where each transform finds good alignment. While this reduced image registration failures, the issue still persisted for quite high proportion of frames, that had to be filtered out. Many of the filtered frames had low light conditions or extreme weather that would be valuable in the training process. In future work, more robust frame alignment methods could be explored for retaining these challenging frames in the training data. 

Dense image registration methods, like Fourier-Mellin that is utilized here are computationally demanding. If the high number of frames need to be processed video stabilization techniques could offer more efficient solution.

Our method transfers a single road annotation to multiple frames relying on the assumption that the road area remains identical. However, snow might accumulate on the road decreasing the size of the drivable area, while our label remains unchanged, causing error to the training data. This issue mostly affects areas close to the road boundaries. 

\section*{Acknowledgment}

We acknowledge the funding provided by Henry Ford Foundation Finland, the computational resources provided by the Aalto Science-IT project, and the data provided by Fintraffic.

\balance
\bibliographystyle{IEEEtran}
\input{main.bbl}


\end{document}